\documentclass[unnumsec,webpdf,contemporary,large, square]{oup-authoring-template}%
\graphicspath{{Fig/}}
\usepackage{graphicx}
\usepackage{epstopdf}

\theoremstyle{thmstyleone}%
\theoremstyle{thmstyletwo}%
\theoremstyle{thmstylethree}%

\hypersetup{colorlinks=true,urlcolor=blue,citecolor=black,linkcolor=black}
\begin{document}

\journaltitle{arXiv}
\copyrightyear{2025}
\pubyear{2019}
\firstpage{1}

%\subtitle{Subject Section}

\title[Recurrent Neural Network for Oncogenic Mutation Progression]{A Novel Recurrent Neural Network Framework for Prediction and Treatment of Oncogenic Mutation Progression}

\author[1,$\ast$]{Rishab Parthasarathy \ORCID{0000-0002-8982-9469}}
\author[2]{Achintya Bhowmik}

\authormark{Parthasarathy and Bhowmik}

\address[1]{\orgname{Massachusetts Institute of Technology}, \orgaddress{\street{77 Massachusetts Avenue}, \postcode{Cambridge}, \state{MA 02139}, \country{USA}}}
\address[2]{\orgname{Stanford University School of Medicine}, \orgaddress{\street{801 Welch Road}, \postcode{Palo Alto}, \state{CA 94304}, \country{USA}}}

\corresp[$\ast$]{Corresponding author: Rishab Parthasarathy, Phone Number: +1 (469)-865-6885, Email: \href{email:email-id.com}{rpartha@mit.edu}}

%\editor{Associate Editor: Name}

%\abstract{
%\textbf{Motivation:} .\\
%\textbf{Results:} .\\
%\textbf{Availability:} .\\
%\textbf{Contact:} \href{name@email.com}{name@email.com}\\
%\textbf{Supplementary information:} Supplementary data are available at \textit{Journal Name}
%online.}

\abstract{Despite significant medical advancements, cancer remains the second leading cause of death, with over 600,000 deaths per year in the US. One emerging field, pathway analysis, is promising but still relies on manually derived wet lab data, which is time-consuming to acquire. This work proposes an efficient, effective end-to-end framework for Artificial Intelligence (AI) based pathway analysis that predicts both cancer severity and mutation progression, thus recommending possible treatments. The proposed technique involves a novel combination of time-series machine learning models and pathway analysis. First, mutation sequences were isolated from The Cancer Genome Atlas (TCGA) Database. Then, a novel preprocessing algorithm was used to filter key mutations by mutation frequency. This data was fed into a Recurrent Neural Network (RNN) that predicted cancer severity. Then, the model probabilistically used the RNN predictions, information from the preprocessing algorithm, and multiple drug-target databases to predict future mutations and recommend possible treatments. This framework achieved robust results and Receiver Operating Characteristic (ROC) curves (a key statistical metric) with accuracies greater than 60\%, similar to existing cancer diagnostics. In addition, preprocessing played an instrumental role in isolating important mutations, demonstrating that each cancer stage studied may contain on the order of a few-hundred key driver mutations, consistent with current research. Heatmaps based on predicted gene frequency were also generated, highlighting key mutations in each cancer. Overall, this work is the first to propose an efficient, cost-effective end-to-end framework for projecting cancer progression and providing possible treatments without relying on expensive, time-consuming wet lab work.}

% \boxedtext{
% \begin{itemize}
% \item Key boxed text here.
% \item Key boxed text here.
% \item Key boxed text here.
% \end{itemize}}

\maketitle

\section{\textbf{Introduction}}\label{intro}
Cancer remains a major challenge for humanity, and despite numerous improvements in treatment over the years, is still the second leading cause of death in the United States, only behind heart disease, with over 600,000 deaths every year \cite{siegel2020}. 

There are three main causes of cancer's continuing challenge. First, complex, late-stage cancers are either often untreatable or develop resistance to treatments such as chemotherapy \cite{housman2014, riggio2020, rawla2019}. Second, at least 25\% of cancer is not caught early, reducing effective treatment outcomes \cite{sung2021}. Third, when signs of precancerous progression are discovered, there is often no way to treat it without surgery \cite{rawla2019}. Thus, early detection and treatment are crucial in saving lives.

\begin{figure}[thpb]
    \centering
    \label{fig:1}
    \includegraphics[width=3.1in]{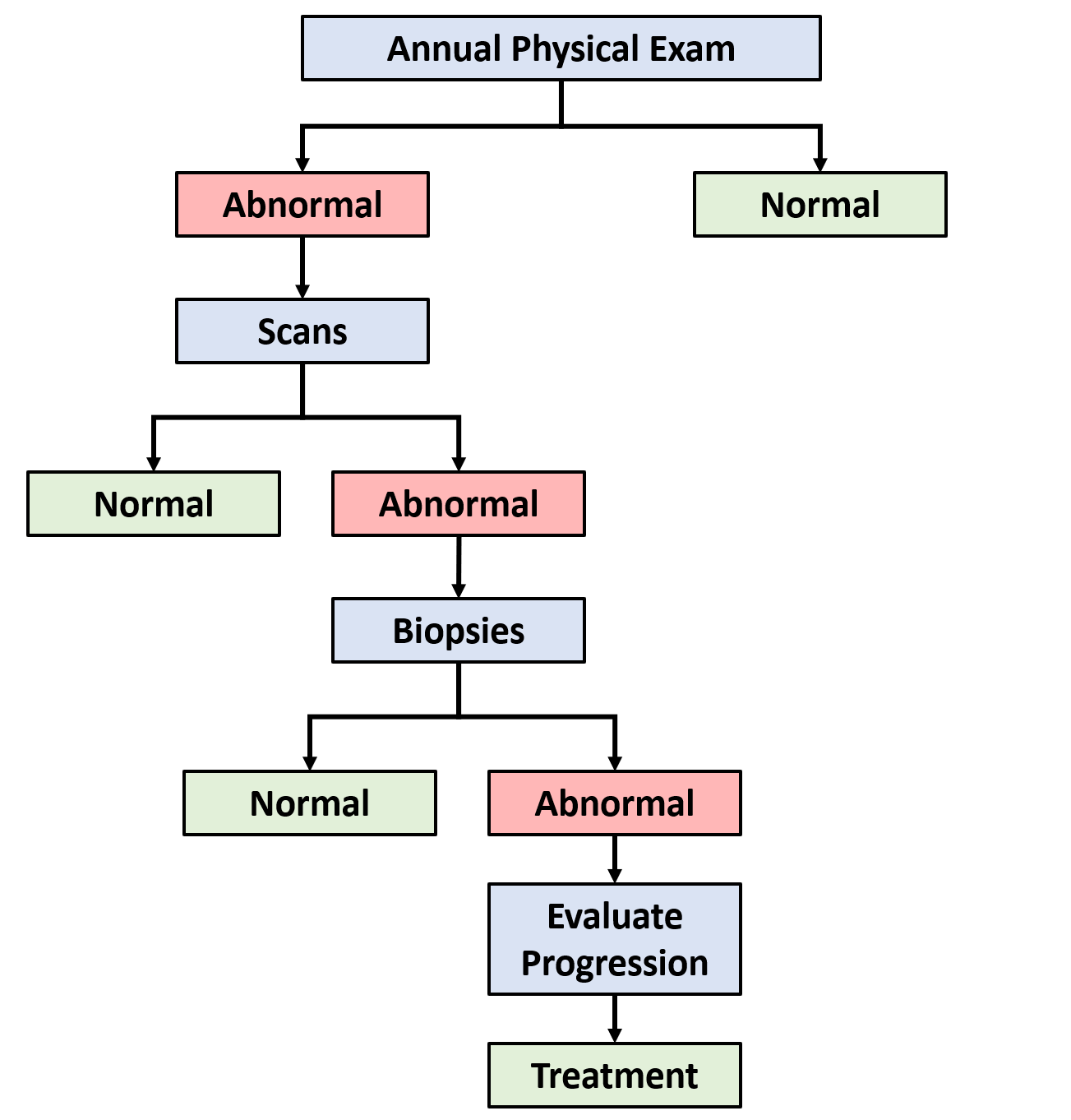}
    \caption{A sample, simplified flow chart that breaks down how oncologists diagnosed cases of thyroid cancer \cite{tonorezos2016}.}
\end{figure}

Currently, doctors use a relatively universal three-step approach to evaluate, diagnose, and treat cancer, starting with annual physical examinations, which determine any abnormalities in the patient's health. If any abnormalities are detected, patients are subjected to a series of scans and biopsies, which allow doctors to localize and identify any possible cancerous lesions. With the knowledge of the cancer, doctors evaluate both the prognosis and progression in order to properly treat the disease outcomes. An example of this paradigm can be found in \hyperref[fig:1]{Fig. 1}, which depicts a simplified workflow of how a group of oncologists diagnosed various cases of thyroid cancer \cite{tonorezos2016}.

However, no straightforward fully-automated mechanism exists for evaluating this complete end-to-end pipeline, with current computational approaches only capable of analyzing scans and biopsies at a fixed point in time \cite{xue2016}.

\section{\textbf{Objectives}}\label{objectives}
In order to better model how doctors diagnose patients, new cancer diagnostic models must evaluate and treat possible disease progression. With the recent advances in genomics and Artificial Intelligence (AI), there are significant opportunities for developing a complete cancer diagnostic framework that can provide more systematic aid to patients.

\begin{figure*}[thpb]
    \centering
    \label{fig:2}
    \includegraphics[width=6.5in]{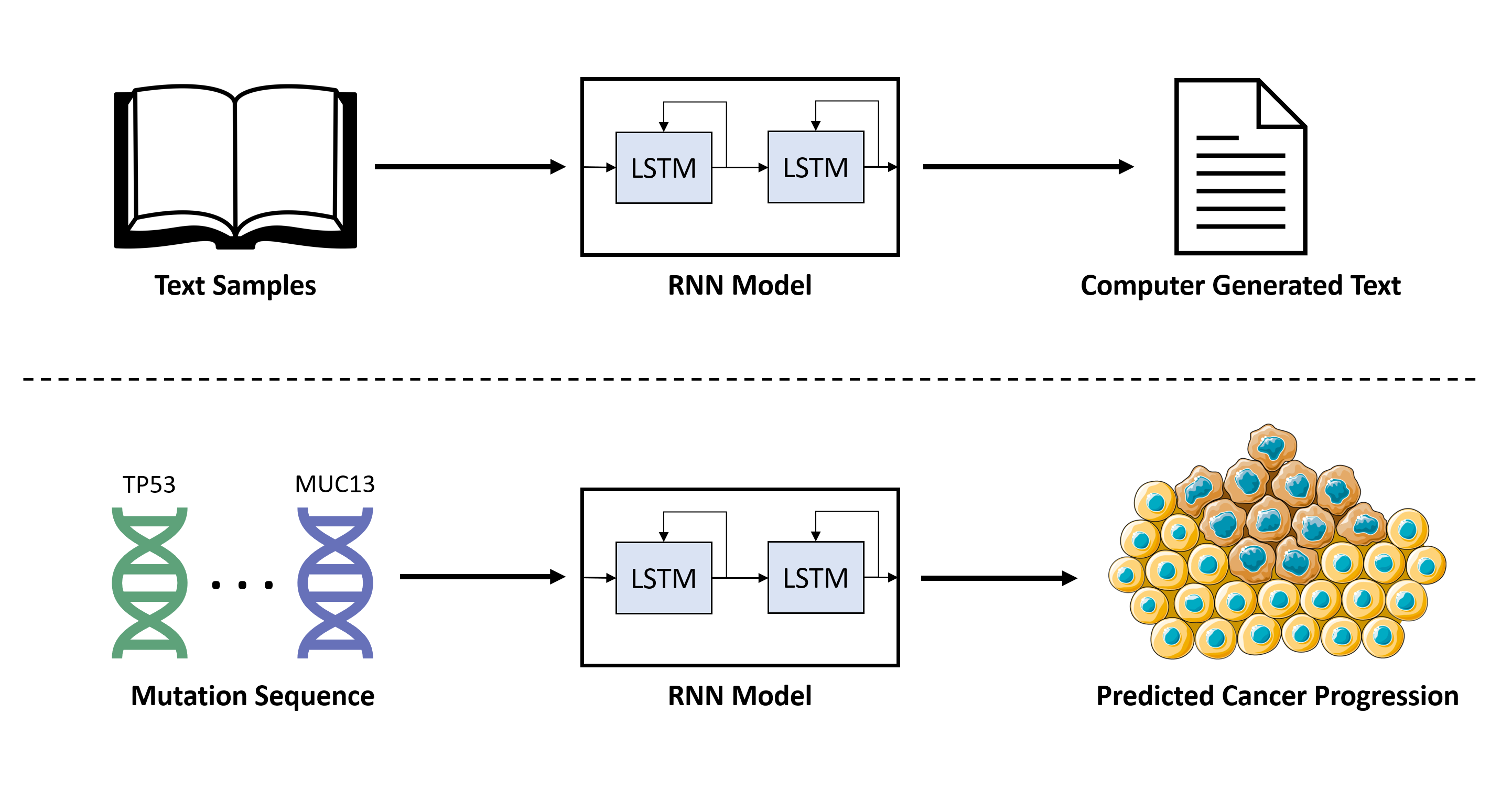}
    \caption{An illustration depicting the parallels between the processing of language and this project's methodology of approaching genomics. In both cases, the input data of text or mutations are fed into an RNN, which learns to infer what will happen in the future through developing spatial and temporal correlations.}
\end{figure*}

This work draws on recent advances in research on time-series processing based on machine learning techniques, specifically the use of Recurrent Neural Networks (RNNs) integrated with Embeddings, which have been validated in contexts from stock market analyses, to most prominently, the Embeddings for Language Models (ELMo) framework for Natural Language Processing (NLP) \cite{karpathy2015,moghar2020}. This work strives to apply the same paradigm to cancer mutation sequences, extracting contextual information from each mutation. In doing so, this work aims to predict not only the present state of cancer, but also the future progression of the disease, possibly unveiling ways to treat cancer symptoms before they even occur. This overall methodology, based on RNN models consisting of Long Short-Term Memory (LSTM) architectures, is portrayed in \hyperref[fig:2]{Fig. 2}.

This work presents an efficient and effective end-to-end framework for machine analysis of biological pathways that will help predict and prevent cancer progression. Using a novel RNN-inspired approach to pathway analysis, this framework provides functionalities for diagnosing cancer, evaluating future cancer progression, and developing targeted drug recommendations using genomic data from a patient's tumor. The goal of this research is to help reduce the burden of cancer on hospitals, doctors, and patients by producing a methodology for targeted treatment of future genomic mutations, demonstrating the feasibility of creating a comprehensive solution for cancer diagnostics.

\section{\textbf{Prior Research}}\label{priors}
\subsection{\textbf{Biological Research}}\label{priorbio}

In the field of bioinformatics, researchers have investigated the use of computational models for analyzing patient scans and biopsies \cite{chougrad2018}. Many approaches have been developed for diagnosing cancer from an image of a Magnetic Resonance Imaging (MRI) scan or photo, mainly focusing on the usage of Convolutional Neural Networks (CNNs), which function by analyzing the spatial correlations within images \cite{chougrad2018,ha2019,jiang2019}. Recent advancements have focused on the use of segmentation models, which allow identification of specific regions of interest, further narrowing analyses \cite{guo2019,kurhousman2014020,mehta2018,isin2016,pereira2017}. However, these methodologies based on feed-forward neural network architectures are not able to provide an analysis of a patient's disease progression as they lack the ability to extract temporal features within time-series data.

Thus, with increasing access to gene sequencing, many researchers have moved towards tackling cancer through genomic analyses \cite{collins2003,berger2019}. 

Automated genomic analyses have focused on using Deep Neural Networks (DNNs). After filtering for relevant genes, these methods feed the genomic data into a series of Fully Connected Layers, which connect all pairs of genes to each other, allowing for large-scale computational calculation \cite{talukder2021,montesinoslopez2021}.

Researchers have also attempted to tackle the genomic aspects of cancer by developing target drugs and gene therapy. Target drugs work by inhibiting a specific gene crucial to a given cancer’s behavior, stopping the cancer in its tracks \cite{sawyers2004}. Treatment with target drugs has already begun to bear fruit, with some late-stage renal cancers becoming curable \cite{ghidini2017}. In gene therapy, faulty genes are replaced or inactivated in order to turn cancerous cells back into normal cells \cite{goncalves2017}. 

However, both these approaches have one key challenge: that it is highly difficult to find exactly what gene to target \cite{buzdin2018}. Gene therapy and target drugs are effective when targeting the correct gene, but often, the incorrect gene is targeted, resulting in ineffectual treatment \cite{buzdin2018,gridelli2011}. These treatments are also expensive, so the overall feasibility of these treatments is still subpar \cite{buzdin2018}.

\begin{figure}[thpb]
    \centering
    \label{fig:3}
    \includegraphics[width=3.2in]{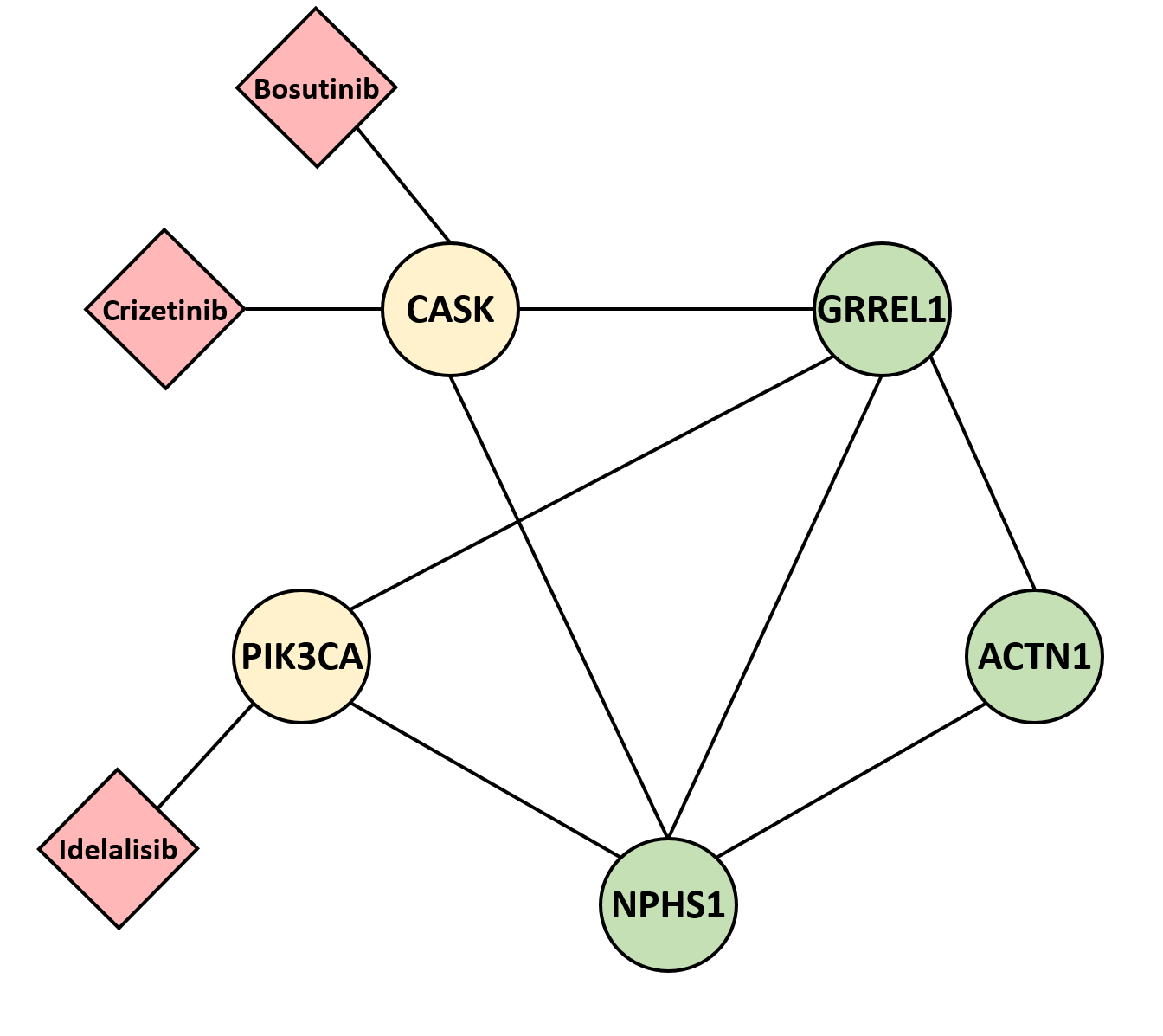}
    \caption{A simplified sample of a snapshot of discovered biological pathways based on manual computation of gene expression in Head and Neck Squamous Cell Carcinoma (HNSCC). Gene-gene interactions are depicted as lines between circles and gene-drug interactions as lines between yellow circles and red squares. These pathways have to be calculated and evaluated by hand in order to verify each one \cite{choonoo2019}.
    }
\end{figure}

One emerging approach for combining these two genomic methodologies is pathway analysis, which analyzes the relationship between genes, gene expressions, and drugs \cite{khatri2012}. The current application of pathway analysis  involves the calculation of coefficients regarding gene interaction or expression in order to determine biological correlations, which are termed ``pathways". These pathways have already proved successful in discovering gene-drug combinations for therapeutic purposes \cite{zolotovskaia2019,yang2020,sivachenko2007}. However, pathway analysis is still limited because it depends on manual processing of wet lab RNA sequencing data in order to verify and determine its discoveries, which is a time-consuming process \cite{khatri2012,zolotovskaia2019}. An example of a simplified snapshot of a pathway analysis framework for Head and Neck Squamous Cell Carcinoma (HNSCC) is presented in \hyperref[fig:3]{Fig. 3}, which depicts gene-gene interactions as lines between circles and gene-drug interactions as lines between yellow circles and red squares \cite{choonoo2019}. 

Specifically, many of these biological relationships require long periods of time to occur, even independent of other biological factors. By integrating the current knowledge of biological pathways with time-series analysis models, this paper aims to derive a new computational methodology for approximating biological pathways through time.

\subsection{\textbf{Time-Series Analysis and Recurrent Neural Networks (RNNs)}}\label{priorrnn}

In recent years, AI research based on time-series analysis techniques has become increasingly prominent through successful applications in fields such as natural language processing, and many of the current models originated from recurrent neural network approaches. The RNN has been used in ubiquitous contexts, from generating Shakespearean plays and language to time-series analyses of the stock market \cite{karpathy2015,moghar2020}. In all these applications, RNN based machine learning architectures have dominated because of their ability to comprehensively generate correlations through time and order \cite{shertinsky2020,karpathy2015,moghar2020}.

Specifically, the RNN architecture wields such power because it embodies the idea of ``attention,'' which is currently used in translation models \cite{shertinsky2020}. Attention is a technique where the existing results from previous time-steps are amplified in order to make more informed decisions at the current time-step. In essence, attention is a way of implementing a more human-like understanding of the context \cite{vaswani2017}. For example, in a language-based context, given the sentence, ``The archer wields a bow," an attention-based model would be able to understand that the word ``bow" means the archer's weapon, not the act of bowing, driven by the context of the word ``archer".

Similarly, mutation sequences often exhibit correlations through time, but despite this apparent connection, RNNs have not yet been comprehensively used for evaluating the progression of cancer mutations \cite{wzhu2020}. Also, this project elects to use the RNN framework over Transformers, another leading attention-based framework because Transformers break down words or lexical units into individual morphemes, which would not help effective training on genomic names and only serve to increase the model complexity and run-time \cite{vaswani2017}. Thus, this project attempts to investigate the parallel between time-series and genomic data by employing RNNs for genomic analyses.

\section{\textbf{Methods}} \label{methods}

\subsection{\textbf{End-to-End Framework}} \label{framework}

\begin{figure*}[thpb]
    \centering
    \label{fig:4}
    \includegraphics[width=6.5in]{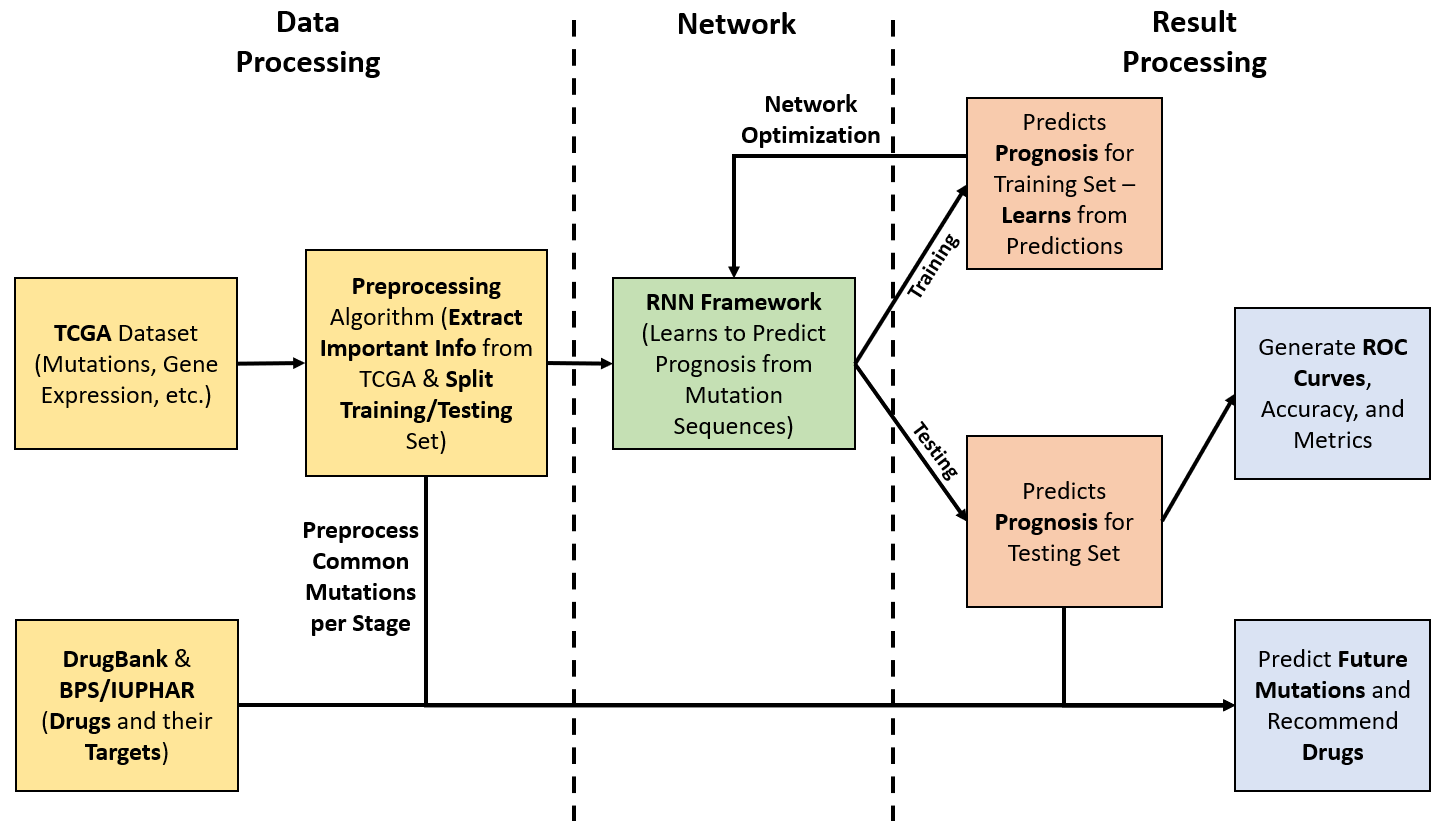}
    \caption{An illustration of the full end-to-end methodology. The Cancer Genome Atlas (TCGA) Dataset was preprocessed in order to find the most salient mutations and split the training/testing set. Then, the RNN framework was trained on the training dataset to accurately predict prognosis. The performance of the RNN was evaluated the testing dataset, generating stage predictions which were used to generate accuracy and Receiver Operating Characteristic (ROC) curves. Finally, the predicted stages, the preprocessed list of important mutations, and the drug databases were used to predict future mutations and drug recommendations.}
\end{figure*}

In this work, a novel methodology for comprehensive analysis of cancer prognosis and progression was developed based on the use of genomic information from patients. As depicted in \hyperref[fig:4]{Figure 4}, there are three phases to the methodology: 1) Data Processing, 2) Network Module, and 3) Result Processing. 

In the Data Processing phase, a preprocessing algorithm was developed to extract the salient information from The Cancer Genome Atlas (TCGA) dataset, filtering for the most common mutations per stage \cite{tcga2022}. After the data was filtered, the Network module, which consisted of an RNN, was trained. Once the model was trained, the RNN predicted the prognosis of the testing data, which was used in combination with information from the preprocessing algorithm in order to predict disease progression and recommend drugs.

\subsection{\textbf{Dataset}} \label{dataset}

In this work, three different datasets were used, all of which performed different purposes, including the training of the neural network and the evaluation of its performance.

The first dataset used was the TCGA dataset, which is the largest open-source genomic dataset on cancer, with the full mutation sequence of more than 20,000 patient samples. The TCGA dataset contains a detailed list of somatic mutations for each patient along with a summary of the patient's type and severity of cancer \cite{tcga2022}. Whenever possible, multiple timepoints for each patient were used; otherwise, cancer stage was used to generate a time-series, as cancer stage represents a linear progression of cancer prognosis through time. For this project, the TCGA dataset was extracted from cBioPortal, an online data repository for cancer genomics \cite{cerami2012,gao2013}.

After extraction from cBioPortal, the classes in the TCGA dataset were evaluated for robustness in training and testing. A hard cutoff of at least 300 samples per class was set, and classes without genomic mutation data were eliminated. As a result, the TCGA dataset used consisted of 11 classes: Bladder Carcinoma (BLCA); Breast Carcinoma (BRCA); Colon Adenocarcinoma (COAD); Head-Neck Squamous Cell Carcinoma (HNSC); Kidney Renal Clear Cell Carcinoma (KIRC); Liver Hepatocellular Carcinoma (LIHC); Lung Adenocarcinoma (LUAD); Lung Squamous Cell Carcinoma (LUSC); Skin Cutaneous Melanoma (SKCM); Stomach Adenocarcinoma (STAD); and Thyroid Carcinoma (THCA) \cite{tcga2022}.

The other two datasets in this project were both used for drug discovery purposes, leveraging existing knowledge of drug-target correlations in order to provide targeted treatment plans. DrugBank, an open-source database run by the University of Alberta, provided the bulk of drug-gene relationships \cite{wishart2018,law2014,knox2011,wishart2008,wishart2006}. In order to ensure the safety and efficacy of the drug treatments discovered, the International Union of Basic and Clinical Pharmacology / British Pharmacological Society (IUPHAR/BPS) Guide to Pharmacology database was used to validate the data in the DrugBank database \cite{harding2021}.

\subsection{\textbf{Data Preprocessing}} \label{preprocess}

To make the TCGA dataset compatible with the RNN framework, a number of preprocessing techniques were applied, which was crucial because of two main challenges. First, many mutations were too rare to have verifiable impacts: for example, in the TCGA BRCA data, only 16.8\% of mutations occurred in more than 1\% of patients (10 patients in total) \cite{tcga2022}. Second, the most expressed mutations were often the most clinically significant: clinical research had already verified that frequently observed mutations such as PIK3CA, TP53, and BRCA1 were key driver mutations in some of the most aggressive, lethal cancers \cite{stratton2009,rajendran2017}.

\begin{figure*}[thpb]
    \label{fig:5}
    \centering
    \includegraphics[width=5in]{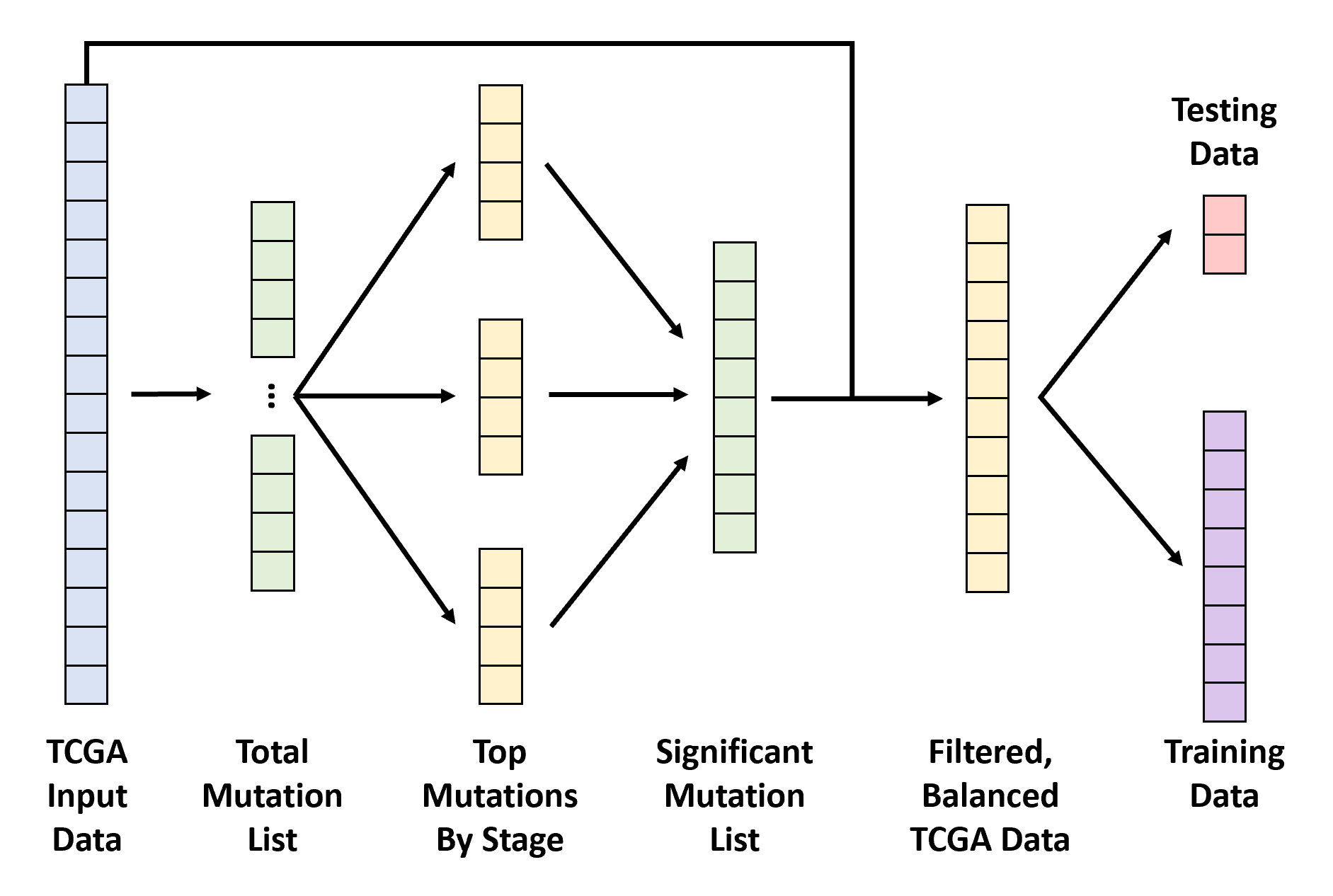}
    \caption{The entire preprocessing paradigm, from the filtering to balancing the class size. The first stage of preprocessing creates a total mutation list from the TCGA input data before calculating the most common mutations both by stage and overall. Then, these commonly observed mutations were used to create a significant mutation list, which was used to filter and balance the TCGA data, which was finally split to a training/testing split.}
\end{figure*}

To preprocess, the algorithm determined the most frequently expressed mutations both overall and in each stage. Based on the expression rates, the algorithm combined the mutation expression list from each stage, creating a list of significant mutations. The algorithm then filtered the TCGA input data to only contain such mutations. Once the data was filtered, the algorithm balanced the class sizes to prevent model overfitting. All in all, this preprocessing method not only simplified the network’s task but also caused increases in the performance as well. The entire preprocessing paradigm is presented in \hyperref[fig:5]{Fig. 5}.

Specifically, each stage which constituted less than 10\% of total data was removed from the data, as there was not enough data to be statistically significant. This modification also helped combat the rapid rate of overfitting inherent to deep neural networks \cite{srivastava2014}.

Then, $S_x$ was calculated as the top $x$ mutations overall, and $S_{x, y}$ was calculated as the top $x$ mutations in stage $y$, sorted by the expression frequency. Using these computed sets, the full mutation list was calculated using \hyperref[eq:1]{Eq. 1}.

\begin{equation}
    \label{eq:1}
    S = \left\{S_x, \dots, \left(S_{x, i} - \left(S_{x, i} \cap \left(S_x \cup \left(\bigcup_{j = 1}^{i - 1} S_{x, j}\right)\right)\right)\right)\right\}
\end{equation}

Once $S$ was computed, the preprocessing algorithm removed all mutations that were not selected from the dataset.

Ultimately, the dataset was balanced by defining a weighted SoftMax transform, depicted in \hyperref[eq:2]{Eq. 2}, where for a sample vector $v$ and weight vector $w$, the output $P$ was calculated by \cite{bridle1989}:

\begin{equation}
    \label{eq:2}
    P_i = \frac{e^{v_i w_i}}{\sum_{j} e^{v_j w_j}}
\end{equation}

To optimize the weighting, as shown in \hyperref[eq:3]{Eq. 3}, the weight vector $w$ was defined using the class sizes $c$ from the data, where

\begin{equation}
    \label{eq:3}
    w_i = \frac{\sum_i c_i}{2 c_i}
\end{equation}

This weighting method prevented overfitting by equalizing the gradients created by each class within the training procedure.

\subsection{\textbf{Recurrent Neural Network (RNN)}} \label{rnn}
 The RNN framework used in this project followed a three-step model that used a sequence of text to generate predictions. In this case, each patient's mutation sequence was used to predict the cancer stage and generate temporal correlations between mutations.

The first step of the RNN was a one-hot embedding, which signified that each mutation was processed as an array of all zeros apart from a single one. The embedding layer then transformed this mutation array into a shorter array of $k$ bounded values. Specifically, this project utilized an embedding of length 256. The mathematical formalism for transforming a one-hot vector $v$ of length $n$ to an embedded vector $e$ of length $k$ is presented in \hyperref[eq:4]{Eq. 4}, given a matrix of weights $w$ \cite{hancock2020}.

\begin{equation}
    \label{eq:4}
    e_j = \sum_{i = 1}^n v_n w_{ij}
\end{equation}

By training the weights, the embedding learned correlations through the similarity between the embedded values. 

The second step of the RNN was a series of Long Short-Term Memory (LSTM) units, which obtained one more piece of information for each time-step (each mutation read) \cite{shertinsky2020}. The LSTMs could then learn temporal correlations in the data, which enabled the prediction of cancer progression.

\begin{figure}[thpb]
    \centering
    \label{fig:6}
    \includegraphics[width=3.2in]{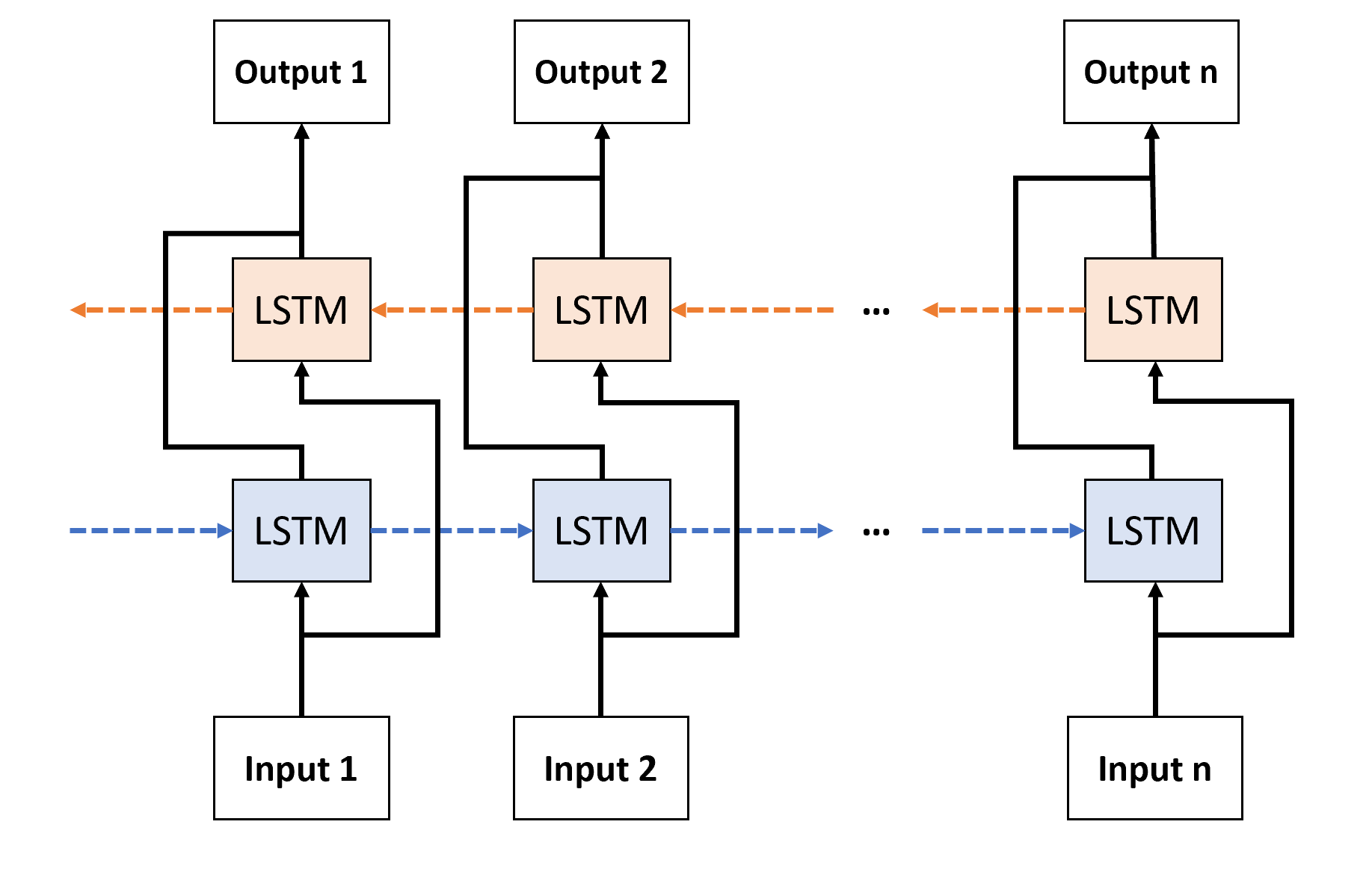}
    \caption{A sample bidirectional LSTM layer, where the blue represents the forward training pass of the algorithm and the orange represents the backward smoothing pass of the algorithm.}
\end{figure}

This project employed a bidirectional LSTM layer, which simultaneously processed the data in both backward and forward directions. The forward pass trained the algorithm while the backward pass smoothed the predictions, allowing more data to be accurately analyzed \cite{shertinsky2020,schuster1997}. A bidirectional LSTM layer is presented in \hyperref[fig:6]{Fig. 6}.

After the LSTM layer, the third and final step of the RNN was a series of Fully Connected, or Dense layers, where each pair of sequential neurons was connected, enabling easy consolidation of information \cite{yamashita2018}. 

\begin{figure*}[thpb]
    \centering
    \label{fig:7}
    \includegraphics[width=6in]{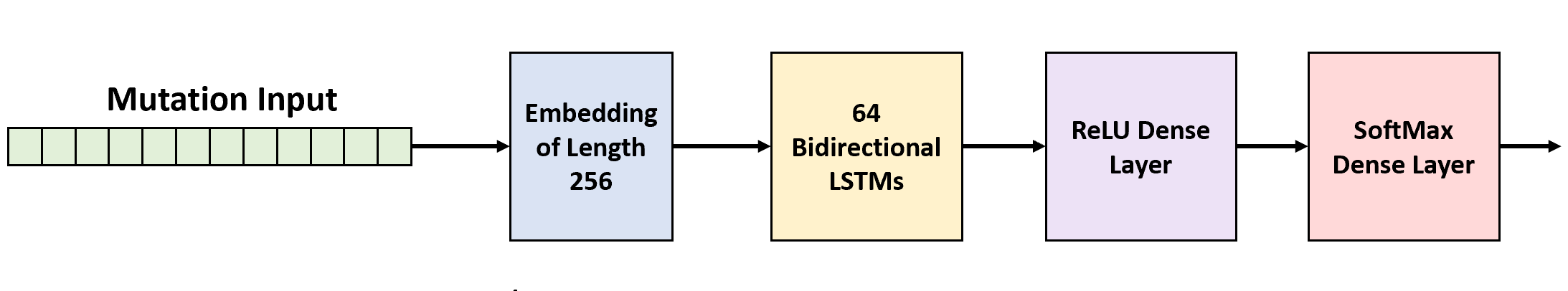}
    \caption{A breakdown of the network structure used, from the Embedding of length 256 and bidirectional LSTM layer of length 64 to the two Dense layers, which were activated with the Rectified Linear Unit (ReLU) and SoftMax, respectively.}
\end{figure*}

Overall, the specific RNN machine learning configuration used in this project contained an Embedding of length 256 (i.e. transforming each mutation into a float matrix of length 256), a bidirectional LSTM layer of length 64, and two Dense layers, which were activated with the Rectified Linear Unit (ReLU) and SoftMax, respectively. A breakdown of this network is presented in \hyperref[fig:7]{Fig. 7}.

\subsection{\textbf{Gene/Drug Prediction}} \label{genedrug}

Once the RNN framework produced a stage prediction, the algorithm then produced future gene predictions and generated drug treatments for those predicted genes.

First, the RNN extracted the mutations that it had correlated with the predicted stage, which were compared against the input mutation list to extract the mutations that had not yet occurred. Through this process, the RNN learned which mutations would occur even months and years into the future.

After extracting these significant future mutations, the postprocessing algorithm calculated the probability of each mutation occurring. This probability was extracted by evaluating the frequency at which each future mutation occurred relative to each input. These probabilities were then used to generate heatmaps, visually portraying the correlation of each mutation to each stage of cancer.

\begin{figure}[thpb]
    \centering
    \label{fig:8}
    \includegraphics[width=3.2in]{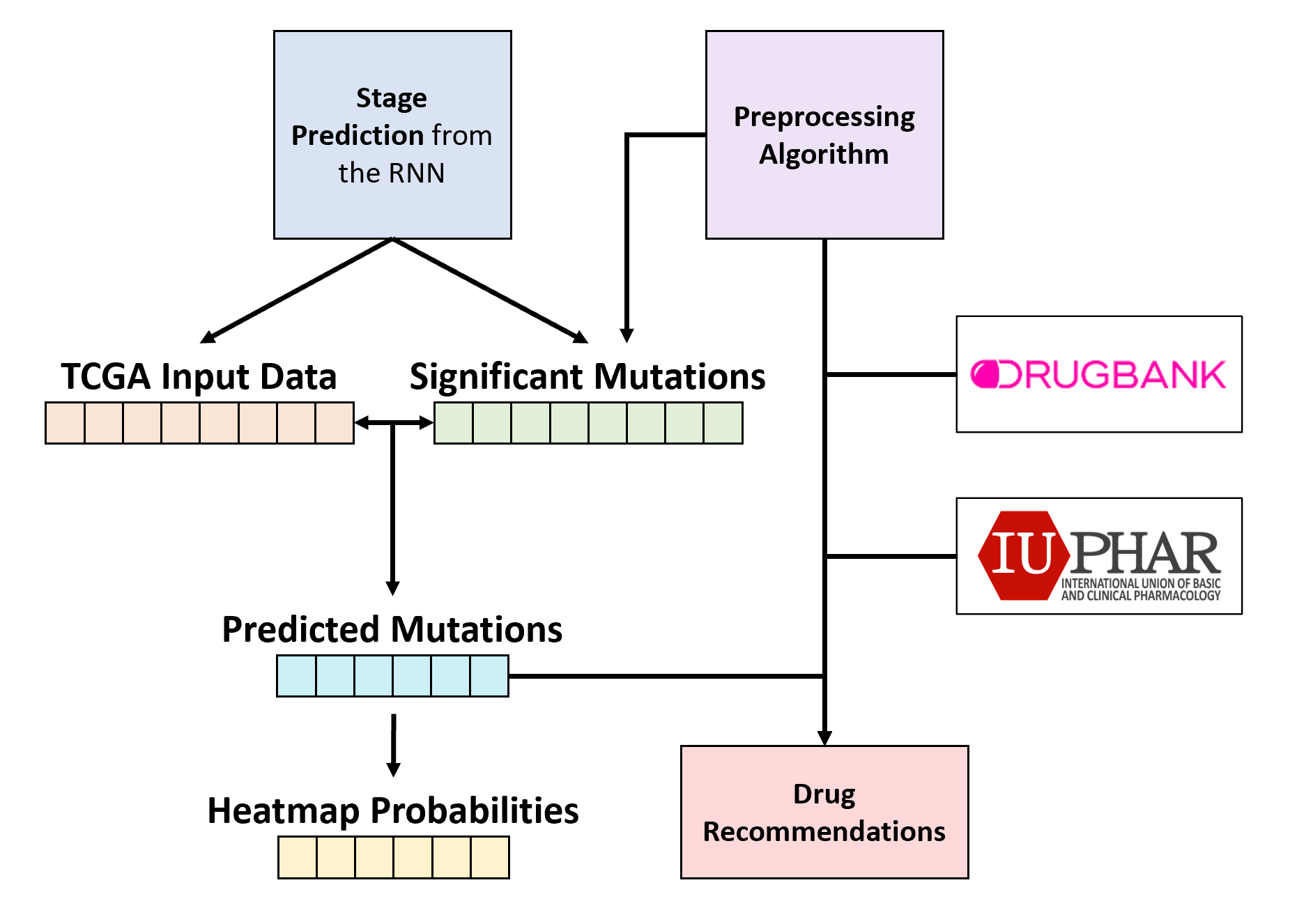}
    \caption{The pipeline used for gene/drug prediction, extracting both heatmaps of significant mutations and providing drug recommendations to treat these mutations even years into the future.}
\end{figure}

In addition, with the driver mutation lists for cancer progression, the algorithm queried the DrugBank and IUPHAR/BPS databases of drug/target interactions, which described how certain drugs modified the behavior of given genes \cite{wishart2018,law2014,knox2011,wishart2008,wishart2006,harding2021}. Using this information, the algorithm  evaluated whether any treatments would treat predicted driver mutations, validating the DrugBank data using the IUPHAR/BPS database. A depiction of this pipeline is provided in \hyperref[fig:8]{Fig. 8}.

\section{\textbf{Results}} \label{results}
Each model was trained for 200 epochs with 80\% of the data assigned to the training set and 20\% assigned to the testing set. Various degrees of preprocessing were utilized in order to validate the effectiveness of the preprocessing algorithm. Specifically, preprocessing for the top 50, 100, and 200 mutations was tested for each cancer.

When relevant, algorithmic performance was evaluated using Receiver Operating Characteristic (ROC) curves, which plot sensitivity against specificity. ROC curves depict the robustness of the algorithm against a purely random output, which is represented by a diagonal line. The performance of ROC curves can be qualitatively evaluated by comparing the curves against the diagonal line (random guessing): a consistent lack of intersection between the curves and the line indicates robustness in the information that the algorithm learned \cite{hajiantilaki2013}.

\subsection{\textbf{Stage Predictions}} \label{stagepred}

\begin{figure*}[thpb]
    \centering
    \label{fig:9}
    \includegraphics[width=6in]{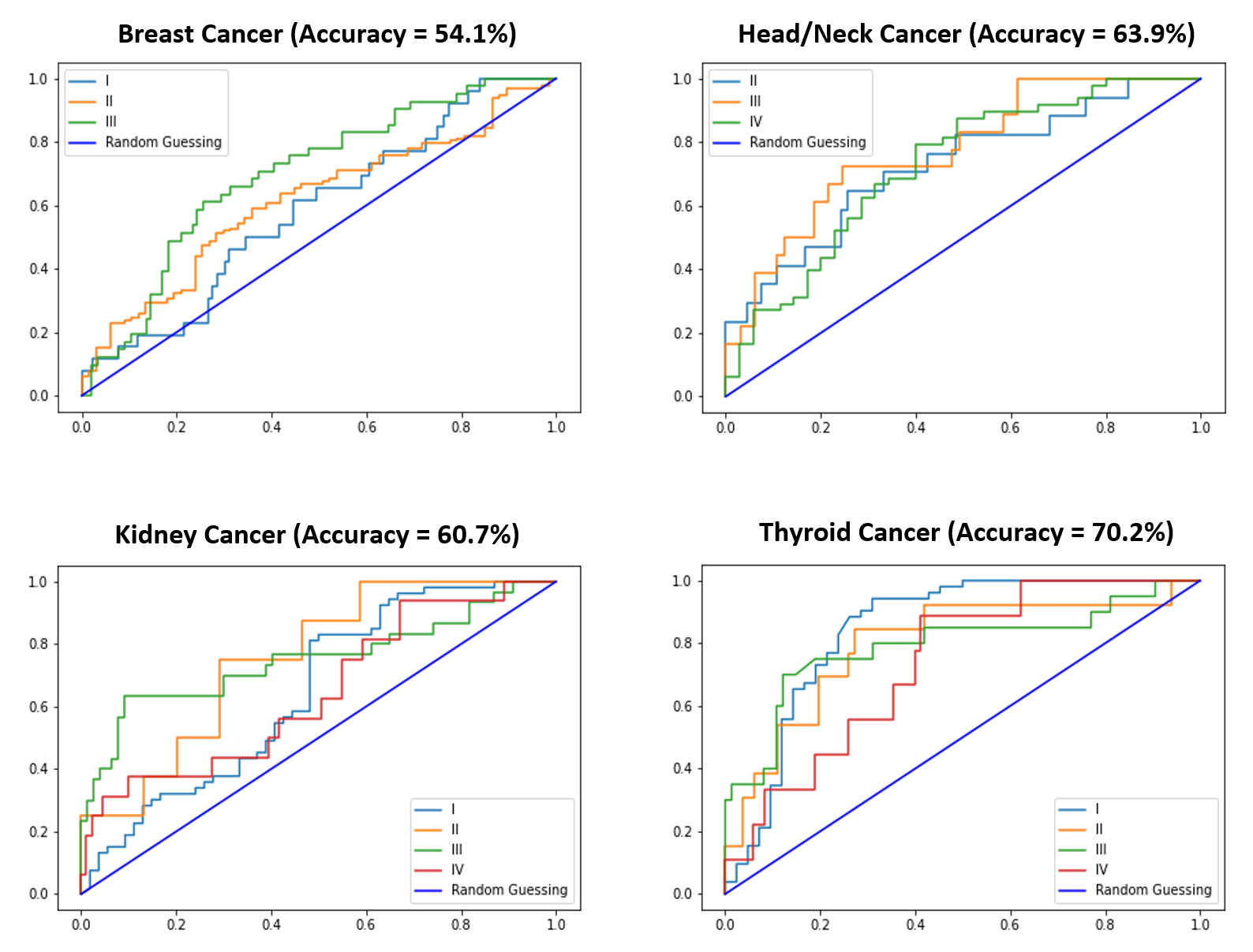}
    \caption{Receiver Operating Characteristic (ROC) curves from four of the cancers that this project evaluated (breast, head/neck, kidney, thyroid). Each ROC curve refers to the cancer stage that was predicted. These ROC curves are all  robust, significantly above Random Guessing, which implies the model’s successful retention of genomic attributes correlated with stage/severity.}
\end{figure*}

To evaluate the effectiveness of the RNN algorithm in predicting cancer stage, the algorithm was run individually on the dataset from each cancer type, and both ROC curves and accuracy were generated. One ROC curve was generated for each cancer stage, and they were grouped by cancer type as presented in \hyperref[fig:9]{Fig. 9}.

For the purpose of clarity, \hyperref[fig:9]{Fig. 9} presents a representative sample of the cancer types tested, distributed throughout different sections of the body. Thyroid cancer represents the endocrine system, kidney cancer the excretory system, head/neck cancer the nervous system, and breast cancer the lymphatic system. 

These results demonstrate that all four models are robust, with ROC curves significantly above the diagonal. In addition, given that no individual ROC curve intersects with the diagonal, the model did not overfit on any specific stage. This behavior confirms the efficacy of the stage weighting procedure used during the preprocessing stage. 

\subsection{\textbf{Preprocessing Performance}} \label{preprocessperf}
\begin{figure*}[thpb]
    \centering
    \label{fig:10}
    \includegraphics[width=6in]{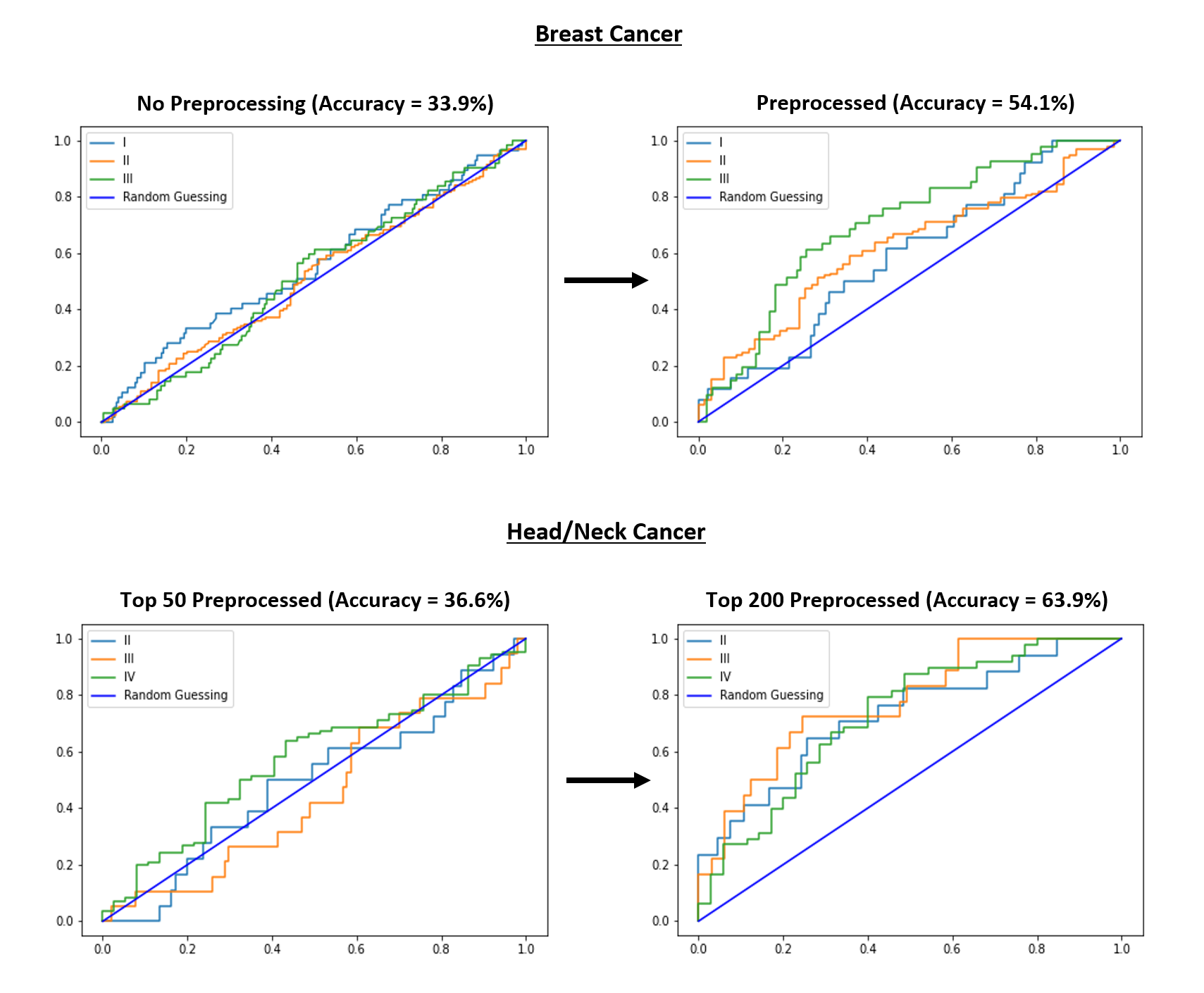}
    \caption{This figure depicts two different insights from the preprocessing algorithm, using representative cancer types. First, preprocessing improves the algorithm performance, as many non-driver mutations are removed, as depicted with breast cancer. Second, preprocessing the top 200 most expressed mutations is most effective for robustness, indicating that there may be on the order of 200 key driver mutations.}
\end{figure*}

The ROC curves also demonstrate important insights from preprocessing, as depicted in the representative examples provided in \hyperref[fig:10]{Fig. 10}. These results clearly indicate that the preprocessing methods enhanced the model's performance, improving from random guessing to true robust predictions, as in the case of breast cancer, with a 1.6-fold increase in accuracy: from 33.9\% to 54.1\%. In addition, the ROC curves demonstrate that by eliminating non-driver mutations, algorithmic performance improved significantly, indicating that the algorithm may not have been able to find long-term correlations from many mutations. However, as with head and neck cancer, preprocessing the top 200 expressed mutations yielded far better results than just 50 mutations, with a 1.75-fold increase between 63.9\% and 36.6\%. This massive increase in accuracy and robustness suggests that there may be on the order of 200 key mutations in head/neck cancer, as there may not have been sufficient information for the model to learn from just 50 mutations. All other cancer types also had optimal performance when preprocessing for 200 mutations compared to 50 mutations and the whole dataset of thousands of mutations, implying that the number of key mutations may be on the order of a few hundred for the types of cancer analyzed in this study.

\subsection{\textbf{Heatmaps}} \label{heatmap}
\begin{figure*}[thpb]
    \centering
    \label{fig:11}
    \includegraphics[width=5in]{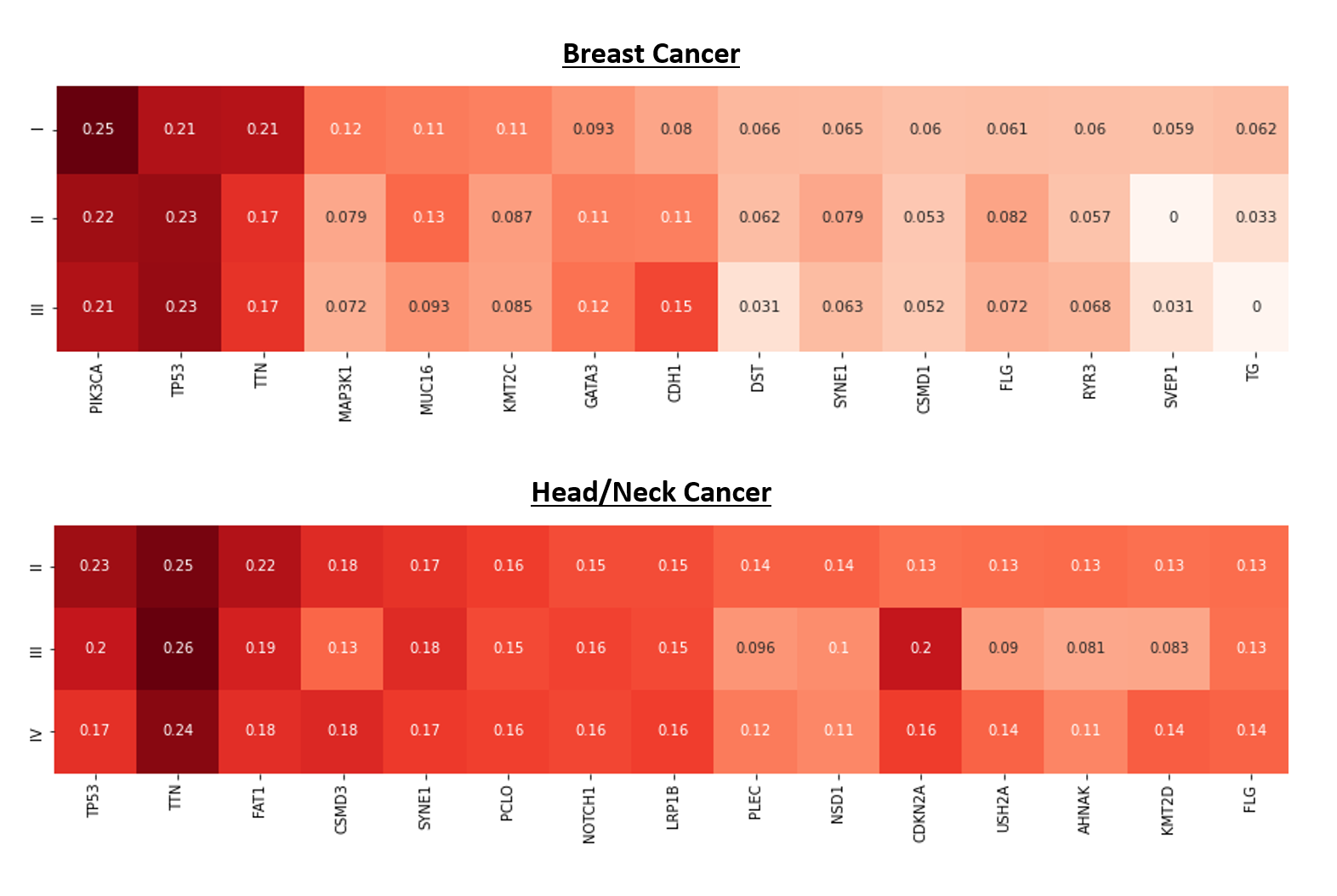}
    \caption{This figure presents two representative heatmaps, correlating cancer stage on the vertical axis to individual cancer mutations on the horizontal axis. Each square represents the summative correlation computed, which is color coded with a darker red indicating a larger correlation.}
\end{figure*}

As shown in \hyperref[fig:11]{Fig. 11}, the heatmaps plot mutation prioritization by stage, with the cancer stage on the vertical axis and the specific gene mutation on the horizontal axis. By learning summative correlations based on gene frequency, the heatmaps facilitate easy identification of key driver mutations per stage. 

For example, PIK3CA and TP53 are indicated as highly correlated with all stages of breast cancer, which is biologically verifiable \cite{martinezsaez2020,chen2022,anderson2020,schon2017,zhu2020,rivlin2011}. In addition, one mutation stands out, CDH1, with a much higher Stage 3 correlation of 0.15 than Stage 1 correlation of 0.08. In fact, CDH1 is also being actively investigated for treatment of aggressive strains of breast cancer, demonstrating the efficacy of the model \cite{pharoah2001,corso2018,corso2016}. Similarly, another key indicator mutation of poor prognosis, CDKN2A, can be extracted from the heatmap for head/neck cancer, as it has experienced a 1.5-fold increase in correlation from Stage 1 to 2 \cite{chen2018,gadhikar2017,zhou2018}. Thus, these heatmaps allow gene prioritization when developing targeted therapies, providing a straightforward approach to evaluating prognostic mutations. 

%\begin{figure*}[thpb]
%    \centering
%    \label{fig:12}
%\includegraphics[width=3in]{Fig12.eps}
%    \caption{This figure depicts a sample drug prediction for a mutation in the gene PIK3CA, generating three viable therapeutic options.}
%\end{figure*}

As for drug predictions, the algorithm predicts drugs based on the specific mutations provided, and one representative example is provided with PIK3CA. For example, the drug prediction generates three possible treatments, alpelisib, copanlisib, and pilaralisib, which are all either in use as key FDA-approved treatments or in highly regarded clinical trials \cite{andre2019,dreyling2019,soria2015}. Once again, the algorithm's predictions are consistent with biological research, demonstrating its effectiveness.

\section{\textbf{Discussion}} \label{discussion}

The framework presented in this paper was capable of computationally predicting and correlating future cancer mutation progression consistent with existing biological data \cite{martinezsaez2020,chen2022,anderson2020,schon2017,zhu2020,rivlin2011,pharoah2001,corso2018,corso2016,chen2018,gadhikar2017,zhou2018,andre2019,dreyling2019,soria2015}. This RNN-based framework had several key advantages over current genomic models. First, by learning from raw data, the model did not require humans to manually parse the input data to discover pathways. In essence, the model functioned without relying on wet-lab RNA-sequencing data, which is time-consuming to produce \cite{khatri2012,zolotovskaia2019}. By predicting mutation progression, this language-inspired algorithm could also help mitigate cancer progression by providing targeted drug treatments, a far more integrated end-to-end framework than existing techniques \cite{xue2016}.

As for preprocessing, this investigation discovered that computational models are most effective when processing the top 200 mutations for each stage, which was observed over all 11 types of cancer investigated. There are two possible reasons for this observation. First, with a high number of mutations, rarely expressed mutations encouraged network overfitting, rendering performance inadequate on sequestered testing datasets. Second, utilization of smaller number of mutations decreased network robustness, implying that there exist key biological pathways specifically encoded within the order of a few hundred driver mutations. This result is consistent with other biological research that has discovered a similar order of a few hundred consistently observed genes with driver mutations \cite{bailey2018,iranzo2018}. 

The model’s accuracy on genomic data then computationally verified a link between mutations and the cancer stage, especially demonstrating the predictive power of utilizing the temporal relationship between mutations. In addition, the prediction accuracy for stage (severity) was either around or greater than 50-60\%, which was comparable to both existing computational models and the performance of medical professionals in estimating cancer prognosis, as presented in \hyperref[table:1]{Table I}, where GAN represents a Generative Adversarial Network, RF represents a Random Forest model, and DNN represents a Deep Neural Network \cite{lopezgarcia2020,kwon2021,malhotra2019}.

\begin{table*}[thpb]
\centering
\label{table:1}
\caption{Comparative Performance of Different Diagnosis Frameworks}
\begin{tabular}{|c|c|c|c|}
\hline
                           & \textbf{Diagnostic Model} & \textbf{Average Accuracy Range} & \textbf{Cancer Types} \\ \hline
This Work                  & RNN                    & 36-70\%   & 11 types              \\ \hline
López-García et al. \cite{lopezgarcia2020} & CNN                    & 68\%  & Lung                  \\ \hline
López-García et al. & ML       & 62-70\% & Lung                \\ \hline
Kwon et al. \cite{kwon2021}         & GAN + CNN              & 41-80\%  & 12 types               \\ \hline
Kwon et al.        & GAN + RF               & 47-74\%  & 12 types               \\ \hline
Kwon et al.        & GAN + DNN              & 42-77\% & 12 types                \\ \hline
Malhotra et al. \cite{malhotra2019}     & Oncologists            & 62\% & Advanced                   \\ \hline
\end{tabular}
\end{table*}

Thus, this model achieved comparable performance to both leading models and a survey of oncologists, suggesting that continued work on this framework may ultimately result in a useful diagnostic and prognostic aid for helping doctors project and treat the progression of a patient's disease.

However, the one outlier in the model's success was its performance on the Colorectal Adenocarcinoma (COADREAD) dataset, on which the model only achieved 36\% accuracy even after preprocessing, as well as Lung Squamous Cell Carcinoma (LUSC) and Skin Cutaneous Melanoma (SKCM), where the model only achieved around 45\% accuracy. Despite being competitive with the numbers proposed by Kwon et al., this may suggest one limitation of the model, that it cannot account for external factors such as lifestyle and environmental circumstances, which can play the most significant roles in causing cancers like melanoma (UV radiation), colorectal adenocarcinoma (diet), and lung squamous cell carcinoma (smoking)  \cite{kwon2021, volkovova2012,parkin2011}. In addition, the TCGA dataset draws from a relatively limited pool of people, so further evaluation on larger, more equitable datasets will be necessary to truly scale this project \cite{spratt2016}.

Despite these limitations, this project serves as a valuable proof-of-concept for RNN-based machine learning approaches to cancer diagnostics, unlocking the possibilities of predicting and preventing mutations before they happen.

\section{\textbf{Conclusion}} \label{conclusion}
Overall, this study was one of the first to apply AI frameworks based on RNN architectures, which are typically used for time-series analysis, to a genomic pathway analysis problem. By proposing, implementing, and evaluating an efficient, cost-effective end-to-end framework, this project demonstrates an RNN-based model for predicting cancer severity, projecting cancer progression, and providing recommendation for possible treatments. In addition, by not relying on formally derived pathway correlations, this project enables rapid computational analysis of genomic data, allowing real-time prognosis prediction and treatment. In doing so, the model presented in this project may enable doctors to better analyze cancer progression, possibly enabling more effective cancer prevention and treatment on a large scale, especially with additional improvements through adversarial training procedures \cite{kwon2021}.

This project has revealed the efficacy of applying a series-analysis based approach to a genomic problem. In the future, analytical methods such as the use of Shapley values may be used to evaluate the internal RNN performance \cite{lundberg2017}. By unveiling the so-called ``black-box" behind the RNN, a continuation of this research may understand the specific techniques and insights that the RNN uses to learn correlations. By combining these computational insights with the existing knowledge of biological pathways, this model may be able to deepen the fundamental understanding of the connection between various genes. In addition, the general paradigm proposed in this project can be extended to other diseases with a genomic correlation, such as cystic fibrosis or Alzheimer’s \cite{rosenberg2016,sharma2020}. Thus, this project serves as a proof-of-concept for an efficient, cost-effective, and generalizable methodology for projecting disease progression and recommending targeted drug treatments, which in the future, could be life-saving in the prevention, diagnosis, and treatment of any genomically-correlated disease, cancer and beyond.

%%%%%%%%%%%%%%

\section{\textbf{Data Availability}}
All data used in this project is publicly available at  \href{https://portal.gdc.cancer.gov/}{\color{blue}{TCGA dataset}}, \href{https://go.drugbank.com/}{\color{blue}{DrugBank database}}, \href{https://www.guidetopharmacology.org/}{\color{blue}{IUPHAR/BPS database}} \cite{tcga2022, cerami2012, gao2013, wishart2018, law2014, knox2011, wishart2008, wishart2006, harding2021}. The codebase developed and used can be found at \url{https://github.com/rishab-partha/Cancer-Progression-Pub}. All derived data is available upon reasonable request.

\section{\textbf{Competing Interests}}
No competing interest is declared.

\section{\textbf{Author Contributions}}
R.P. and A.B. both contributed to and edited the manuscript. R.P. designed the study and performed the methods and analyses, with A.B. serving in an advisory role. Both R.P. and A.B. reviewed the paper.

\section{\textbf{Key Points}}
\begin{itemize}
    \item Recurrent Neural Network (RNN)-based Artificial Intelligence (AI) frameworks allow for the simultaneous modeling of cancer severity and mutation progression, as demonstrated in this work.
    \item Using data from The Cancer Genome Atlas (TCGA) dataset, preprocessing algorithms were used in conjunction with an RNN framework to diagnose cancer stage and identify key cancer driver mutations implicated in cancer progression.
    \item Cancer mutation predictions can be used in conjunction with drug-target databases to provide preemptive drug recommendations for patients with evolving cancers.
    \item Simple RNN-based frameworks are competitive with other more complex frameworks in the task of cancer severity, while also pairing greater generalizability in analyzing correlations.
    \item With analysis of environmental data and larger datasets, RNN-based frameworks possess potential for extension to other types of cancer as well as other time-correlated diseases.
\end{itemize}

\section{\textbf{Funding}}
This work was not funded by any external sources.

\section{\textbf{Acknowledgments}}
The authors thank the anonymous reviewers for their valuable suggestions. The authors also thank the teachers at the Harker School, especially Mr. Chris Spenner and Dr. Eric Nelson for their contributions to reviewing early versions of this paper.

\bibliographystyle{unsrt}
%\bibliography{reference}

%USE THE BELOW OPTIONS IN CASE YOU NEED AUTHOR YEAR FORMAT.
%\bibliographystyle{abbrvnat}
%\bibliography{reference}

%% sample for biography with author's image

%% sample for biography without author's image
\begin{biography}{}{\author{Rishab Parthasarathy} is a student at the Massachusetts Institute of Technology interested in the practical application of Electrical Engineering and Computer Science, especially in the application of Artificial Intelligence to medicine/health. Rishab is a published author in IEEE journals and was awarded a scholarship as a 2022 Top 40 Finalist in the Regeneron Science Talent Search. He has also earned an International Physics Olympiad Gold Medal and International Linguistics Olympiad Silver Medal.}
\end{biography}

\begin{biography}{}{\author{Achintya Bhowmik} is an adjunct professor at the Stanford University School of Medicine and the Wu Tsai Neurosciences Institute, where he advises research and lectures in the areas of sensory augmentation, computational perception, and intelligent systems. He is the chief technology officer and executive vice president of engineering at Starkey, a privately-held medical devices company. He is an elected Fellow of the Institute of Electrical and Electronics Engineers (IEEE), the Asia-Pacific Artificial Intelligence Association (AAIA), and the Society for Information Display (SID). He has authored over 200 publications, including two books and over 80 granted patents.}
\end{biography}

\end{document}